\documentclass{elsarticle}

\usepackage{graphicx}

\usepackage{amssymb}

\usepackage{lineno}
\modulolinenumbers[5]

\usepackage{booktabs}
\usepackage{enumitem}
\usepackage{longtable}
\usepackage{graphicx}
\usepackage{array}
\usepackage{dcolumn}
\usepackage{amsfonts}
\usepackage{amssymb}
\usepackage{xspace}
\usepackage{xfrac}
\usepackage{array}
\usepackage{multirow}
\usepackage{xcolor}
\usepackage{footnote}
\usepackage{hyperref}
\usepackage{makeidx}
\usepackage{epsfig}
\usepackage{amsmath}
\usepackage{multicol}
\usepackage{mathtools}
\usepackage[misc]{ifsym}
\usepackage[ruled,linesnumbered]{algorithm2e}

\begin{document}

\begin{frontmatter}

\title{Model Compression for Dynamic Forecast Combination}

\author[label1]{Vitor Cerqueira\corref{cor1}}
\ead{cerqueira.vitormanuel@gmail.com}

\author[label1]{Luis Torgo}
\ead{ltorgo@dal.ca}

\author[label2,label3,label4]{Carlos Soares}
\ead{csoares@fe.up.pt}

\author[label5,label6]{Albert Bifet}
\ead{abifet@waikato.ac.nz}

\address[label1]{Faculty of Computer Science, Dalhousie University, 6050 University Ave, Halifax, NS B3H 1W5, Canada}
\address[label2]{Fraunhofer AICOS Portugal, Porto, Portugal}
\address[label3]{INESC TEC, Porto, Portugal}
\address[label4]{University of Porto, Porto, Portugal}
\address[label5]{Télécom ParisTech, Paris, France}
\address[label6]{University of Waikato, Hamilton, New Zealand}

\cortext[cor1]{Corresponding author}

\begin{abstract}

The predictive advantage of combining several different predictive models is widely accepted. Particularly in time series forecasting problems, this combination is often dynamic to cope with potential non-stationary sources of variation present in the data.
Despite their superior predictive performance, ensemble methods entail two main limitations: high computational costs and lack of transparency. These issues often preclude the deployment of such approaches, in favour of simpler yet more efficient and reliable ones. 
In this paper, we leverage the idea of model compression to address this problem in time series forecasting tasks.
Model compression approaches have been mostly unexplored for forecasting. Their application in time series is challenging due to the evolving nature of the data. Further, while the literature focuses on neural networks, we apply model compression to distinct types of methods. 
In an extensive set of experiments, we show that compressing dynamic forecasting ensembles into an individual model leads to a comparable predictive performance and a drastic reduction in computational costs. Further, the compressed individual model with best average rank is a rule-based regression model. Thus, model compression also leads to benefits in terms of model interpretability.
The experiments carried in this paper are fully reproducible.

\end{abstract}

\begin{keyword}
forecasting \sep ensemble methods \sep model compression \sep interpretability \sep rule-based regression

\end{keyword}

\end{frontmatter}


\section{Introduction}\label{sec:introduction}

\subsection{Context}

Ensemble methods methods combine the output of different models to solve a given predictive task. These approaches have been shown to provide state of the art predictive performance in many domains of application. 
This advantage shows up in many studies, both empirical and theoretical \citep{brown2005managing}. 
Further, this benefit is usually attributed to the reduction of the risk of selecting a poor model \citep{hibon2005combine}.

In time series forecasting problems, ensemble methods have attracted much attention for a long time~\citep{clemen1989combining}.
The combination process is typically dynamic, which means that the weights of each model in the available pool change over time. The rationale behind a dynamic approach is to cope with the different dynamic regimes governing the time series. Accordingly, at a given point in time, each model is weighted according to their expected predictive performance in the upcoming observations.

Despite being the benchmark approach in terms of predictive performance, ensemble methods are often not applicable in practice due to two issues: computational requirements \citep{bucilua2006model} and lack of interpretability \citep{lakkaraju2016interpretable}. 

Ensembles are computationally expensive in terms of time and space. 
Learning ensemble scales linearly with the number of individual predictors, which may lead to prohibitive computational costs when the number of models is high. 
In this scenario, the costs of storing the models is also relevant.
These issues are especially deployment is in small edge devices, which is increasingly relevant in embedded systems related to the \textit{internet-of-things}. 
Moreover, when time series data is collected in high frequency scenarios (e.g. data streams), the time necessary to execute all models in the ensemble is often prohibitive. 
The temporal constraint is magnified when the ensembles are dynamic, which is almost always the case in time-dependent data, because the weights of each model need to be frequently updated. 

On top of these computational problems, ensembles are hard to interpret \citep{lakkaraju2016interpretable}. Transparent models, which enable the understanding of the mechanisms behind their decisions, are essential for trustworthiness and respective adoption by practitioners. 
Additionally, transparency of predictive models is becoming increasingly regulated by authorities.

Motivated by the computational problem,  \cite{bucilua2006model} presented the idea of model compression. 
Model compression works by having a single predictive model (student), which is trained to emulate the behaviour of an ensemble (teacher). Henceforth, we will refer to this approach as student--teacher (ST) training paradigm. This approach differs from the \textit{traditional} way of learning a predictive model by replacing the original target variable with the predictions of an ensemble. Here, the ensemble works as a predictive model with a highly flexible functional form. 

The idea of model compression gained considerable attention when \cite{hinton2015distilling} applied the ST paradigm using ensembles of deep neural networks to automatic speech recognition problems at Google. Interestingly, the authors manage to apply an ST training procedure and obtain a single, more compact, deep neural network with a comparable predictive performance relative to the ensemble.

\subsection{Our Approach}

Current ST training approaches are mostly based on compressing large ensembles of deep neural networks. Moreover, these address classification tasks.
The objective of our work is to study the application of model compression to forecasting numeric time series problems. This scenario is particularly relevant for two main reasons:
\begin{itemize}
    \item Time series data typically changes over time due to the presence of concept drift \citep{gama2014survey}.
    This is an additional challenge for the process of compressing a dynamic ensemble;
    
    \item Time series prediction in resource-limited environments is an important research topic in the machine learning literature. The successful application of the ST training approach potentially enables the deployment of strong yet compact predictive models in small devices.
\end{itemize}

We apply model compression to tackle univariate time series forecasting tasks using dynamic heterogeneous ensembles. We build a portfolio of predictive models using different regression learning algorithms to predict the future behaviour of the time series. The models in this portfolio are dynamically combined at run-time to cope with the non-stationary sources of variation present in the data. 
We performed experiments using 90 data sets from several domains of application. For each problem, we combined a set of 30 predictive models using 14 distinct combination approaches, including stacking \citep{wolpert1992stacked}, arbitrage \citep{cerqueira2019arbitrage}, and several dynamic approaches from the online learning literature \citep{Cesa-Bianchi:2006:PLG:1137817}. The results suggest that:
\begin{enumerate}[label=(\roman*)]
    \item Student models (individual models trained using the ST approach) have a competitive predictive performance with dynamic ensembles; 
    
    \item Training individual models using the ST approach leads to a significantly better performance relative to training them using the original data.
\end{enumerate}

\noindent We remark that, while performing comparably, student models require significantly lower computational resources relative to dynamic ensembles. 
The best student model is a variant of the model tree algorithm \citep{quinlan1993combining}, which is trained to emulate the behaviour of Stacking \citep{wolpert1992stacked}. 
Therefore, the application of model compression leads to benefits in terms of model interpretability \citep{liu2018improving}, which is a desirable property for model deployment \citep{lakkaraju2016interpretable}. 

Our contributions can be summarised as follows:
\begin{itemize}
    \item A novel approach for time series forecasting tasks which is based on the compression of \emph{dynamic} \emph{heterogeneous} ensembles into a single individual model. To our knowledge, this is the first work in the literature to apply model compression methods to time series forecasting tasks. Moreover, while most model compression approaches focus on neural networks, our proposal is agnostic to the learning algorithms applied;
    
    \item An extensive set of experiments which validate the proposal. The results from these experiments show: (i) the competitiveness of the individual compressed model relative to the dynamic ensembles in terms of predictive performance; a significant gain in terms of computational costs.
\end{itemize}

\noindent We believe that our research can have a considerable impact on forecasting systems. Particularly in resource-aware environments, for example, sensors devices related to the internet-of-things or sensitive applications which require model transparency.

Besides the introduction, this paper is organized with five more sections. In the next section, we overview the related work. In Section \ref{sec:methodology}, we describe the model compression methodology, and how we apply it to time series prediction. In Section \ref{sec:experiments}, we present an extensive set of experiments which validate the application of the ST training approach for forecasting. Finally, we conclude the paper in Section \ref{sec:discussion}.

In order to encourage reproducible research, the R code and data sets used in the experiments is available online\footnote{\url{https://github.com/vcerqueira/model_compression_forecasting}}.

\section{Related Research}

In this section, we will provide some background on ensemble methods, list the state of the art approaches for combining predictive models, and summarise some of the critics pointed at this type of approaches.
We will give particular emphasis to ensemble approaches designed for time series forecasting problems. 
Finally, we will also overview related work in model compression and knowledge distillation. 

\subsection{Ensemble Methods}

Over the years, many learning algorithms have been developed, from decision trees to support vector machines. However, it is widely accepted that no particular method is best suited for all problems. 
This idea is supported by the \textit{No Free Lunch} theorem for supervised learning \citep{wolpert1996lack}.
The hypothesis that every particular predictive model has some limitations is the primary motivation behind ensemble methods \citep{brown2005managing}. 

The goal of ensemble methods is to combine the predictions of different predictive models. Diversity among the individual models is known to be a crucial component in ensemble methods \citep{brown2005managing}. 
In other words, different models should provide overall good but different predictions from one another. 

In summary, the idea is to have different models that are better at different parts of the data space to manage the limitations of each one.

\subsection{Dynamic and Heterogeneous Ensembles for Forecasting}\label{sec:fcapproaches}

Ensemble methods have been applied to many domains, including forecasting \citep{bates1969combination}. \cite{hibon2005combine} argues that combining different forecasting models essentially reduces the risk of selecting the wrong one. 

Time series data often comprises different dynamic regimes. This leads to a phenomenon where different forecasting models show varying relative performance \citep{aiolfi2006persistence}. 
Dynamic ensemble methods fit naturally in this scheme. Initially, a portfolio of models is built, in which these models follow distinct assumptions regarding the underlying process generating the time series observations. 
The idea is that this approach introduces a natural diversity in the ensemble, which is helps to handle the dynamic regimes. 
Ensembles comprised of individual models with \textit{distinct assumptions}, or different inductive biases, are said to be heterogeneous \citep{kuncheva2004classifier}.
One of the main challenges when applying ensemble methods on time series data is determining which hypothesis, or set of hypotheses, is stronger in a given point in time. In evolving environments such as time series, at is common to adopt dynamic methods, where the weights of each model in the ensemble vary over time to cope with concept drift. We will overview some dynamic methods used to combine a set of forecasting models. We split these into three dimensions: windowing approaches, regret minimisation approaches, and meta-learning approaches.

\subsubsection{Windowing Approaches}\label{bg:windowing}

The simple average of the available forecasting models (equal weights) is a robust combination method \citep{clemen1986combining} (referred to here as \texttt{Simple}). Simple averages are sometimes complemented with model selection before aggregation, also known as trimmed means (\texttt{SimpleTrim}). For example, \citep{jose2008simple} propose trimming a percentage of the worst forecasters in past data and average the output of the remaining experts.

A common way to dynamically weight forecasting models is using predictive performance in a window of recent data. The idea behind this approach is that the short-term future will be similar to recent past and earlier observations are not as relevant. Windowing approaches have been used to weight and combine the available models \citep{newbold1974experience,cerqueira2017dynamic} (\texttt{WL}), or to select the recent best performing one \citep{Blast} (\texttt{BLAST}).

Similarly to windowing, one can use a forgetting factor to give more importance to recent values. The adaptive ensemble combination (\texttt{AEC}) follows this approach using an exponential re-weighting strategy to combine forecasting models.

\subsubsection{Regret Minimisation}

In the online learning literature \citep{Cesa-Bianchi:2006:PLG:1137817}, several approaches have been proposed to aggregate the predictions of a set of forecasts. Many of these approaches offer theoretical guarantees in terms of regret, which can be defined as the average error suffered with respect to the best we could have obtained. For a thorough read, we refer the reader to the second chapter of the seminal work by \cite{Cesa-Bianchi:2006:PLG:1137817}.
In this work, we will focus on the following approaches:  the exponentially weighted average (\texttt{EWA}), the polynomially weighted average (\texttt{MLpol}), and the fixed share aggregation (\texttt{FS}). \cite{zinkevich2003online} proposed a method based on online gradient descent (\texttt{OGD}) for minimising dynamic regret.

\subsubsection{Combining by Learning}\label{bg:metal}

Meta-learning is also commonly used for forecast combination. In this case, we refer to meta-learning methods as those that apply learning procedures to estimate the weights of each model in the ensemble. \texttt{Stacking} is a widely used approach which falls within this category \citep{wolpert1992stacked}. Another approach is the application of ridge regression (\texttt{Ridge}) to the predictions of the forecasting models \citep{gaillard2015forecasting}. According to this approach, the predictions are linearly combined, but their weights do not necessarily sum to one.

\cite{cerqueira2019arbitrage} proposed a meta-learning approach called Arbitrated Dynamic Ensemble (\texttt{ADE}) to combine a set of forecasting models dynamically. 
The gist of the idea is, given a new example, weight each forecasting model according to the prediction of the error they are expected to incur.
A meta-learning layer of models produces the predictions of error.

\subsection{Drawbacks of Ensemble Methods}\label{sec:drawbacks}

\subsubsection{Computational Costs}

Despite being the benchmark in terms of predictive performance in many tasks, including forecasting, ensembles learning approaches are not always viable in practice.
Ensembles scale linearly with the number of models that compose them, which can quickly lead to unacceptable computational costs
As we described in the introduction, this issue is due to space and time constraints. The first relates to the memory necessary to store the predictive models composing the ensemble, while the latter concerns the time necessary to retrieve the predictions of each model and combine them to make a final decision. 

The problems outlined above are particularly relevant in time-dependent tasks. For example, in data stream mining, predictive models need to be able to process observations with limited temporal and spatial resources.

\subsubsection{Interpretability}

Optimizing predictive performance, which ensemble methods excel at, is sometimes not enough for deploying a predictive model \citep{lakkaraju2016interpretable}.
In some domains of application, it is fundamental to build transparent models, which enable the understanding of the process from the input of a new observation to the output decision.
Ensemble methods typically lack transparency \citep{guidotti2018survey}. Even if the models that constitute the ensemble are transparent, this interpretability is often diluted when aggregating these models.

\subsection{Compacting Predictive Models}

Compacting or accelerating predictive models has received a lot of attention in recent years. In a survey, \cite{cheng2017survey} split the approaches to solve this task into four categories: (i) Parameter pruning or sharing, in which obsolete parameters are removed -- pruning decision trees fall into this category; (ii) Low-rank factorization, which involves decomposing tensors for estimating important parameters; (iii) Transferred convolutional filters, in which a set of base filters are used to build the convolutional layers in neural networks; and (iv) model compression (or knowledge distillation). 
In this work, we focus on model compression approaches to reduce the computational complexity of dynamic ensembles. 
To the best of our knowledge, this is the first approach to model compression in time series forecasting. We have focused on dynamic ensembles, as these are extensively studied to address time series forecasting tasks and represent the state of the art approach in terms of predictive performance. 

\subsubsection{Model Compression}\label{sec:mc}

Model compression, presented by \cite{bucilua2006model}, was designed to overcome the computational costs of ensemble methods. The idea is to train a model, designated as a student, to mimic the behaviour of an ensemble (the teacher). First, the idea is to retrieve the predictions of the teacher in observations not used for training (e.g. a validation data set). Then, the student model is trained using this set of observations, where the explanatory variables are the original ones, but the original target variable is replaced with the predictions of the teacher.

\cite{bucilua2006model} use the ensemble selection algorithm \citep{caruana2004ensemble} as the teacher and a neural network as the student model and address eight binary classification problems. Their results show that the compressed neural network performs comparably with the teacher while being ``1000 times smaller and 1000 times faster''. Moreover, the compressed neural network considerably outperforms the best individual model in the ensemble used as the teacher.

\cite{hinton2015distilling} developed the idea of model compression further, denoting their compression technique as distillation. Distillation works by softening the probability distribution over classes in the softmax output layer of a neural network.
The authors address an automatic speech recognition problem by distilling an ensemble of deep neural networks into a single and smaller deep neural network. Similarly to \cite{bucilua2006model}, 
their results show that the distilled network obtained results comparable to the ensemble it learned from.
Further, the distilled network significantly outperforms a deep neural network of the same size but trained using the original data.

\subsubsection{Further Developments to Model Compression}

After the seminal works by \cite{bucilua2006model} and \cite{hinton2015distilling}, several developments to model compression have been proposed in the literature.

\cite{zhang2018deep} presented the idea of mutual learning with neural networks. Mutual learning denotes a knowledge distillation process without the need of a powerful teacher. Instead, an ensemble of neural network students learn collaboratively from each other. Empirical results show the competitiveness of mutual learning relative to standard distillation \citep{hinton2015distilling}.

\cite{romero2014fitnets} showed that model compression can also be used to mimic the internal representations of a teacher neural network. On a similar note, \cite{luo2016face} carried out model compression using the neurons of the higher hidden layer of the teacher network. According to the authors, these units are as informative as the soft output probabilities, but more compact.

Model compression can also be applied in the parallel training of deep neural networks \citep{sun2017ensemble}. In order to achieve a compact final model, the authors employ a loss function (denoted as accelerated compression loss) which encourages compression.

These works are related to ours in the sense that they apply model compression to compact a large model into a small one. However, our work is distinguished from previous ones in two important aspects:
\begin{itemize}
    \item We address a numerical prediction problem with non-i.i.d. data. On the other hand, although the soft labels used during compression are numeric, the literature of model compression is focused on classification tasks;
    
    \item Our approach is agnostic to the underlying learning algorithm. The literature is focused on compressing neural network teachers (either ensembles of neural networks or a neural network with a large number of parameters). The student is also often a neural network, though we found some works which use tree-based models. These are described in the next section. Overall, while neural networks have shown a competitive performance in many predictive tasks, ensemble methods composed of different learners are commonly used for time series forecasting tasks \citep{de2014follow,gaillard2015forecasting,montero2020fforma}.
\end{itemize}

\subsubsection{Improving Interpretability with Model Compression}

Compressing an ensemble into an individual model also improves interpretability.
This matter was noted by \cite{che2016interpretable} and developed by \cite{liu2018improving}. They address classification problems in the healthcare domain, in which model transparency is critical. In order to retain interpretability and maximize predictive performance, they compress a deep neural network into tree-based algorithms. The resulting model retains a competitive performance while being more compact and transparent.

Our goal is to apply model compression to ensembles designed for time series forecasting. In resource-aware time series, predictive models have temporal and/or spatial constraints. In this context, compressing a dynamic ensemble may lead to more compact models without sacrificing too much predictive performance.

\section{Model Compression for Forecasting}\label{sec:methodology}

In this section, we formalise the application of model compression to dynamic heterogeneous ensembles designed for tackling univariate time series forecasting tasks.

\subsection{Problem Definition}

In this work, we focus on univariate time series forecasting problems. A time series represents a temporal sequence of values $Y = \{y_1, y_2, \dots,$ $y_n \}$, where $y_i \in \mathbb{R}$ is the value of $Y$ at time $i$ and $n$ is the length of $Y$. Time series forecasting denotes the task of predicting the next value of the time series, $y_{n+1}$, given the previous observations of $Y$. We focus on an auto-regressive modelling approach, predicting future values of time series using its past $p$ lagged values. 

Using common terminology in the machine learning literature, a set of observations ($x_i$, $y_i$) is constructed. In each observation, the value of $y_i$ is modelled based on the past $p$ values before it: $x_i = \{y_{i-1}, y_{i-2}, \dots, y_{i-p} \}$, where $y_i \in \mathbb{Y} \subset \mathbb{R}$, which represents the vector of values we want to predict, and $x_i \in \mathbb{X} \subset \mathbb{R}^p$ represents the feature vector. In summary, the time series is transformed into the data set $\mathcal{D}(x,y) = \{x_i, y_i\}^{n-p+1}_{1}$.
The objective is to construct a model for approximating $f : \mathbb{X} \rightarrow \mathbb{Y}$, where $f$ denotes the regression function. In other words, the principle behind this approach is to model the conditional distribution of the $i$-th value of the time series given its $p$ past values: $f$($y_{i} | x_i$). In essence, this approach leads to a multiple regression problem. The temporal dependency is modelled by having past observations as explanatory variables.

In this paper, we focus on univariate and one-step ahead time series forecasting problems. Notwithstanding, the methodology is applicable to multivariate time series and multi-step ahead forecasting problems.

The methodology we describe for model compression in forecasting settles on two main steps: 
\begin{enumerate}[label=(\roman*)]
    \item Creating an ensemble, which is the \textit{teacher} model;
    
    \item Training the \textit{student} model using the predictions from the teacher. This model is deployed as a compact version of the teacher.
\end{enumerate}

We will describe these two steps in the following two sub-sections.

\subsection{Creating an Ensemble}

Let $\mathcal{D}_{tr}(x,y)$ denote a set of training observations and $\mathcal{D}_{ts}(x,y)$ a set of testing observations\footnote{When appropriate, we also use the subscripts $tr$ or $ts$ when referring to each component individually. For example, $x_{tr}$ denotes the feature set of training observations}. 
As outlined before, the learning objective is to approximate the true unknown function $f$. Our approach is based on ensemble methods. We construct a portfolio of distinct models $\mathcal{H} = \{h_1, h_2, \dots, h_m\}$, where $h_i$ is the \textit{i}-th model in the ensemble. The predictions of each individual model ($\hat{h}$) are aggregated together to make a final decision, denoted as $\hat{H}$. This aggregation can be formalized as follows for a given \textit{j}-th observation.

\begin{equation}\label{eq:agg}
    \hat{H} = \sum_{i=1}^{m} w^j_i\hat{h}^j_i
\end{equation}

\noindent where $w^j_i$ denotes the weight of the \textit{i}-th model in the \textit{j}-th observation.

We focus on heterogeneous ensembles. The set $\mathcal{H}$ of predictive models is created in parallel and independently from each other. 
The use of different learning algorithms to train each model $h \in \mathcal{H}$ promotes diversity, which is essential for ensemble learning.

Estimating the weighting factors of each model in the ensemble is a challenging task. In time-dependent data, the relative performance of each model varies and the respective weight needs to be updated accordingly. Hence, ensembles are typically dynamic. In Section \ref{sec:fcapproaches}, we overviewed several methods used to this effect.

\subsection{Compressing an Ensemble}

We discussed in Section \ref{sec:mc} that, despite their typical superior predictive performance, ensemble methods often require too much computational resources. \cite{bucilua2006model} proposed the idea of model compression to cope with this problem. In this section, we describe the application of model compression to forecasting ensembles.

\begin{figure}[t]
\centering
\includegraphics[width=\textwidth, trim=0cm 3cm 2cm 0cm]{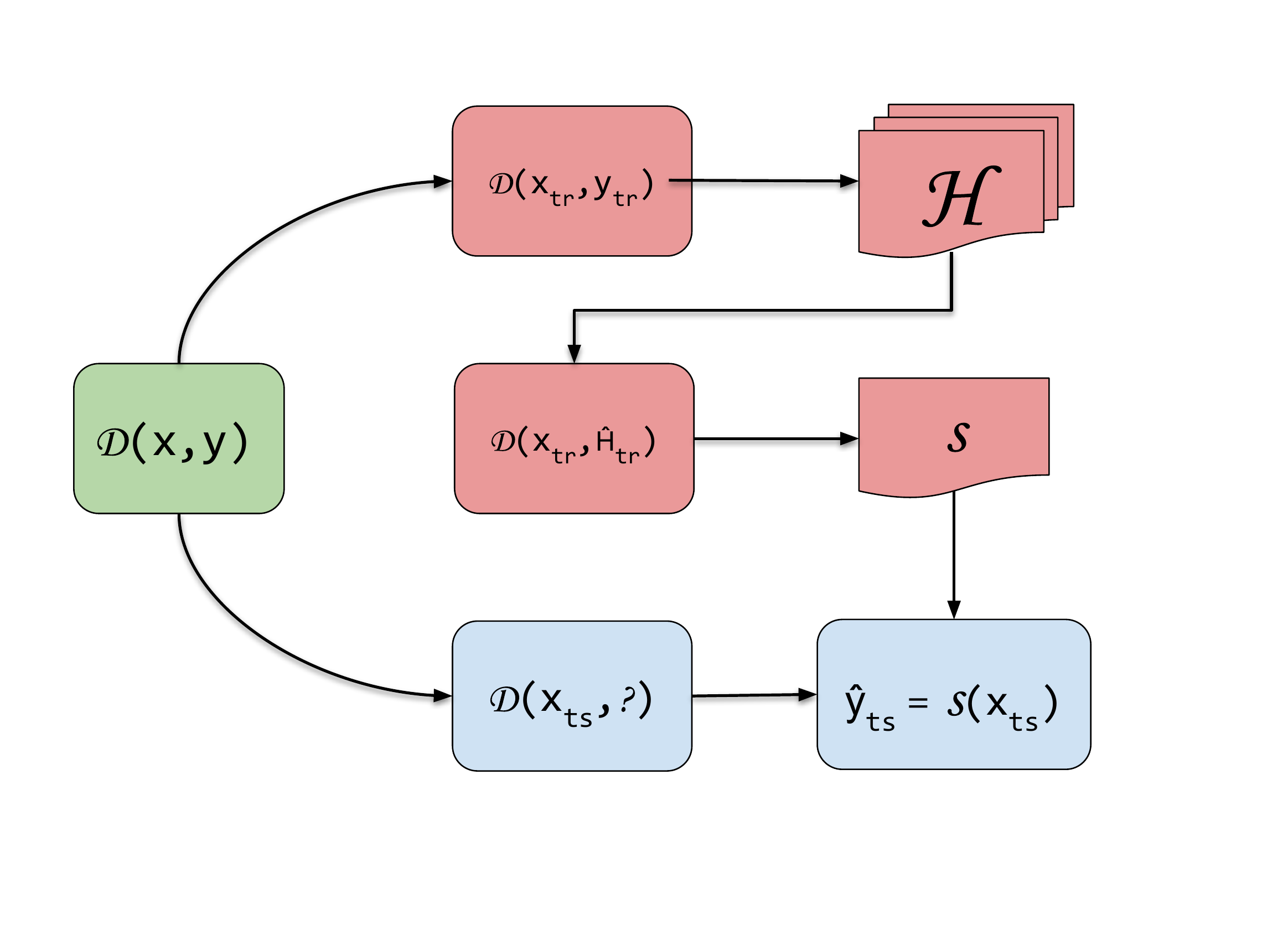}
\caption{Workflow of model compression for forecasting}
\label{fig:workflow}
\end{figure}

The workflow for applying model compression to forecasting ensembles is depicted in Figure \ref{fig:workflow}. Following this illustration, this workflow can be described as follows:

\begin{enumerate}
    \item The available data set is partitioned into training and testing sets, which are denoted as $\mathcal{D}_{tr}(x,y)$ and $\mathcal{D}_{ts}(x,y)$, respectively;
    
    \item The set of predictive models $\mathcal{H} = \{h_1, h_2, \dots, h_m\}$ is trained using $\mathcal{D}_{tr}(x,y)$; 
    
    \item In turn, the ensemble $\mathcal{H}$ is used to make predictions on the training observations $x_{tr}$. The combined predictions of the ensemble are denoted as $\hat{H}_{tr}$;
    
    \item The data set $\mathcal{D}_{tr}(x,\hat{H})$, i.e., the original feature set but replacing the original target variable with the predictions of the ensemble, is used to train the student model (denoted as $s$). We will discuss the use of training observations in Section \ref{sec:teachingdata};
    
    \item Finally, at run-time, the student model $s$ is used to make predictions on the test set. In this phase, the teacher model is discarded.
\end{enumerate}

\subsubsection{Teaching Data}\label{sec:teachingdata}

\cite{bucilua2006model} refer that one of the main challenges behind the application of the ST training approach is the need for unlabeled data. That is, the dataset which the ensemble method labels and is used to train the student model. In the case of forecasting, \textit{labeling} denotes the process of predicting the next value of a time series given its recent past. 
\cite{bucilua2006model} overcome this problem by creating synthetic examples using a method called MUNGE. This method is also formalized by the authors, and we refer to the respective work for a more detailed read \cite{bucilua2006model}. 
\cite{cerqueira2019arbitrage} deal with a similar problem when retrieving data for training a meta-learning layer of models for time series forecasting. They deal with the problem using a blocked prequential approach. Basically, they repeat several learning plus testing cycles using contiguous blocks of observations to obtain unbiased predictions from the training data. 
Notwithstanding, \cite{hinton2015distilling} suggest that using the original training set works well for model compression. We use this approach in the formalization in the last section. In the next section, we will compare this approach with the blocked prequential one used by \cite{cerqueira2019arbitrage}.

\section{Empirical Evaluation}\label{sec:experiments}

In this section, we provide an extensive set of experiments which show the applicability of model compression to forecasting ensembles.

\subsection{Research Questions}\label{sec:rq}

The experiments were designed to address the following research questions:

\begin{description}
    \item[\textbf{Q1}:] How does a dynamic ensemble designed for time series forecasting perform relative to an individual model trained to emulate its behaviour?
    
    \item[\textbf{Q2}:] How does the predictive performance of an individual model trained with the original target variable compare with that of the same individual model trained with the predictions of an ensemble?
    
    \item[\textbf{Q3}:] Following \textbf{Q2}, which student learning algorithm presents the best predictive performance?
    
    \item[\textbf{Q4}:] How does the predictive performance of the above-mentioned approaches compare with that of traditional forecasting methods?
    
    \item[\textbf{Q5}:] Are out-of-bag predictions necessary for training the student models or can this be accomplished using predictions from the training set?
    
    \item[\textbf{Q6}:] What is the relative computational gain by compressing of an ensemble into a single individual model?
    
\end{description}

\subsection{Methods}

The portfolio $\mathcal{H}$ used in the experiments is composed by 30 individual models ($m = 30$). These were created using the following learning algorithms: support vector regression (\texttt{SVR}) \citep{kernlab04}, multivariate adaptive regression splines \citep{mars2016} (\texttt{MARS}), random forest \citep{ranger2015} (\texttt{RF}), projection pursuit regression \citep{R} (\texttt{PPR}), rule-based regression \citep{Cubist2014} (\texttt{RBR}), multi-layer perceptron \citep{monmlp} (\texttt{MLP}), generalised linear model regression \citep{glmnet2010} (\texttt{GLM}), gaussian processes \citep{kernlab04} (\texttt{GP}), principal components regression \citep{plspackage} (\texttt{PCR}), and partial least squares \citep{plspackage} (\texttt{PLS}). Table \ref{tab:expertsspecs} describes the different parameters used for each predictive model.

\begin{table}[!bth]
	\centering
	\caption{Summary of the parameters of the learning algorithms}		
	\resizebox{.9\textwidth}{!}{%
	\begin{tabular}{llll}
	\toprule
	\textbf{ID} & \textbf{Algorithm} & \textbf{Parameter(s)} & \textbf{Value(s)}\\
	\midrule
	    \texttt{SVR\_1} & \multirow{3}{*}{Support Vector Regr.} & \multirow{3}{*}{Kernel} & Radial\\
	    
	    \texttt{SVR\_2} &  &  & Laplace\\
	    
	    \texttt{SVR\_3} &  &  & Polynomial\\
	    
	    \midrule   
	    
	    \texttt{MARS\_1} & \multirow{6}{*}{Multivar. A. R. Splines} & \multirow{6}{*}{\{Degree, No. terms\}} & \{1, 10\} \\
	    
	    \texttt{MARS\_2} & & & \{1, 20\} \\
	    
	    \texttt{MARS\_3} & & & \{1, 30\} \\
	    
	    \texttt{MARS\_4} & & & \{3, 10\} \\
	    
	    \texttt{MARS\_5} & & & \{3, 20\} \\
	    
	    \texttt{MARS\_6} & & & \{3, 30\} \\
	    
	    \midrule   
	    
	    \texttt{RF\_1} & \multirow{2}{*}{Random forest} & \multirow{2}{*}{\{No. trees, mtry\}} & \{500, 5\} \\
	    
	    \texttt{RF\_2} & & & \{500, 10\} \\
        
	    \midrule 
	    
	    \texttt{PPR\_1} & \multirow{6}{*}{Proj. pursuit regr.} & \multirow{6}{*}{\{No. terms, Method\}} & \{2, super smoother\}\\
	    
	    \texttt{PPR\_2} & & & \{4, super smoother\}\\
	    
	    \texttt{PPR\_3} & & & \{6, super smoother\}\\
	    
	    \texttt{PPR\_4} & & & \{2, spline\}\\
	    
	    \texttt{PPR\_5} & & & \{4, spline\}\\
	    
	    \texttt{PPR\_6} & & & \{6, spline\}\\
	    
	    \midrule 
        
        \texttt{RBR\_1} & \multirow{2}{*}{Rule-based regr.} & \multirow{2}{*}{No. iterations units} & 1 \\
        
        \texttt{RBR\_2} & &  & 25 \\
        
        \midrule   
        
        \texttt{MLP\_1} & \multirow{3}{*}{Multi-layer Perceptron} & \multirow{3}{*}{Hidden units} & 5 \\
        
        \texttt{MLP\_2} & & & 7 \\
        
        \texttt{MLP\_3} & & & 10 \\
	    
	    \midrule   
	    
	    \texttt{GP\_1} & \multirow{3}{*}{Gaussian Processes} & \multirow{3}{*}{Kernel} & Radial \\
	    
        \texttt{GP\_2} &  &  & Laplace \\
        
        \texttt{GP\_3} &  &  & Polynomial \\
        
        \midrule
        
        \texttt{PCR} & Principal Comp. Regr. &  No. components & \textit{Auto} \\
        
        \midrule   
        
        \texttt{PLS} & Partial Least Regr. & Method & \{SIMPLS\} \\
        
		\bottomrule    
	\end{tabular}%
	}
	\label{tab:expertsspecs}
\end{table}

In terms of forecast combination methods, we focus on the following approaches.

\begin{description}[leftmargin=*]
    \item[\texttt{Stacking}:] An adaptation of stacking \citep{wolpert1992stacked} for times series forecasting. We apply this approach according to \cite{cerqueira2019arbitrage}. Essentially, the data set used to train the meta-model comprised the predictions of each model in the ensemble. These predictions are obtained according to a blocked prequential approach;
    
    \item[\texttt{Ridge}:] Dynamic aggregation approach based on ridge regression. The predictions of the available methods are linearly combined, but the combination is not necessarily convex;
    
    \item[\texttt{Simple}:] Combination of the predictions of each model according to the arithmetic mean;
    
    \item[\texttt{SimpleTrim}:] A variant of \texttt{Simple}, in which only half of the best past performing available models are combined;
    
    \item[\texttt{WL}:] Adaptive weighted average of the predictions of the available models \citep{newbold1974experience}. The weights are computed according to the predictive performance in the last $\lambda$ observations;
    
    \item[\texttt{BLAST}:] A variant of \texttt{WL} \citep{Blast}, which selects the model with best predictive performance in the last $\lambda$ observations;
    
    \item[\texttt{AEC}:] The adaptive ensemble combination (\texttt{AEC}) procedure \citep{sanchez2008adaptive}, which employs an exponential re-weighting strategy to combine the available models;
    
    \item[\texttt{EWA}:] The exponentially weighted average, which is based on the Weighted Majority Algorithm \citep{Cesa-Bianchi:2006:PLG:1137817};
    
    \item[\texttt{FS}:] The fixed share approach, which is designed to cope with non-stationary time series;
    
    \item[\texttt{MLpol}:] The polynomially weighted average combination approach;
    
     \item[\texttt{OGD}:] A dynamic combination based on online gradient descent, which is designed to minimize a metric for dynamic regret;
    
    \item[\texttt{Best}:] A baseline which select the individual model in the ensemble with best performance in the training data to predict all the test instances.
    
\end{description}

Most of these combination approaches are dynamic to cope with the non-stationarities present in the time series. The exceptions are \texttt{Stacking}, \texttt{Simple} and \texttt{Best}.

The implementation of the methods \texttt{OGD}, \texttt{MLpol}, \texttt{FS}, \texttt{EWA}, and \texttt{Ridge} was from the the R package \textit{opera} \citep{opera}.
In order to augment the data used to estimate the weights, we applied a blocked prequential procedure using the training data \citep{cerqueira2019arbitrage}.

On top of these, we also compare the performance of the methods mentioned above with traditional forecasting approaches, namely \texttt{ARIMA}, \texttt{ETS}, and \texttt{TBATS}. We apply these methods according to their automatic implementation from the \textit{forecast} R package \citep{forecast}.

\subsection{Datasets and Experimental Setup}\label{sec:es}

We centre our study in univariate time series. 
We use a set of time series from the benchmark database \textit{tsdl} \citep{tsdlpackage}. From this database, we selected all the univariate time series with at least 1000 observations and which have no missing values. This query returned 55 time series. These show a varying sampling frequency (daily, monthly, etc.), and are  from different domains of application (e.g. healthcare, physics, economics). For a complete description of these time series we refer to the database source \citep{tsdlpackage}. We also included 35 time series used by \cite{cerqueira2019arbitrage}. Essentially, from the set of 62 used by the authors, we selected those with at least 1000 observations and which were not originally from the \textit{tsdl} database (since these were already retrieved as described above). We refer to the work by \cite{cerqueira2019arbitrage} for a description of the time series.

In summary, our analysis is based on 90 time series. We set the embedding size ($p$) to 15. Notwithstanding, this parameter can be optimised using, for example, the False Nearest Neighbours method or by looking at the auto-correlation plots of the time series. We set the $\lambda$ parameter (the window size used to estimate predictive performance) to 50, based on the results by \cite{cerqueira2019arbitrage}.



Regarding evaluation, we estimate the predictive performance using a holdout applied in multiple testing periods. We sampled at random 10 points in the time series. For each point, we repeated a learning plus testing cycle. In each cycle, the 60\% of the total observations before the sampled point are used for training. The subsequent 10\% observations are used for testing. Repeating the holdout procedure in multiple random testing periods has been shown to provide more robust performance estimates relative to other approaches \citep{cerqueira2019evaluating}.
We use the mean absolute scaled error (MASE) as evaluation metric, which is typically used to evaluate the quality of forecasting models \citep{hyndman2006another}. Essentially, MASE denotes the mean absolute error of a given model scaled by the mean absolute error of a benchmark model. The benchmark model is the naive method, which predicts the next value of a time series to be the value of last known observation. Notwithstanding, in most of our analysis we will compare different forecasting approaches according to their rank. A method with rank of 1 in a time series means that that method is the best performing one (lowest MASE) in that task. 

\subsection{Exploratory Analysis of Results}

\begin{figure}[t]
\centering
\includegraphics[width=\textwidth]{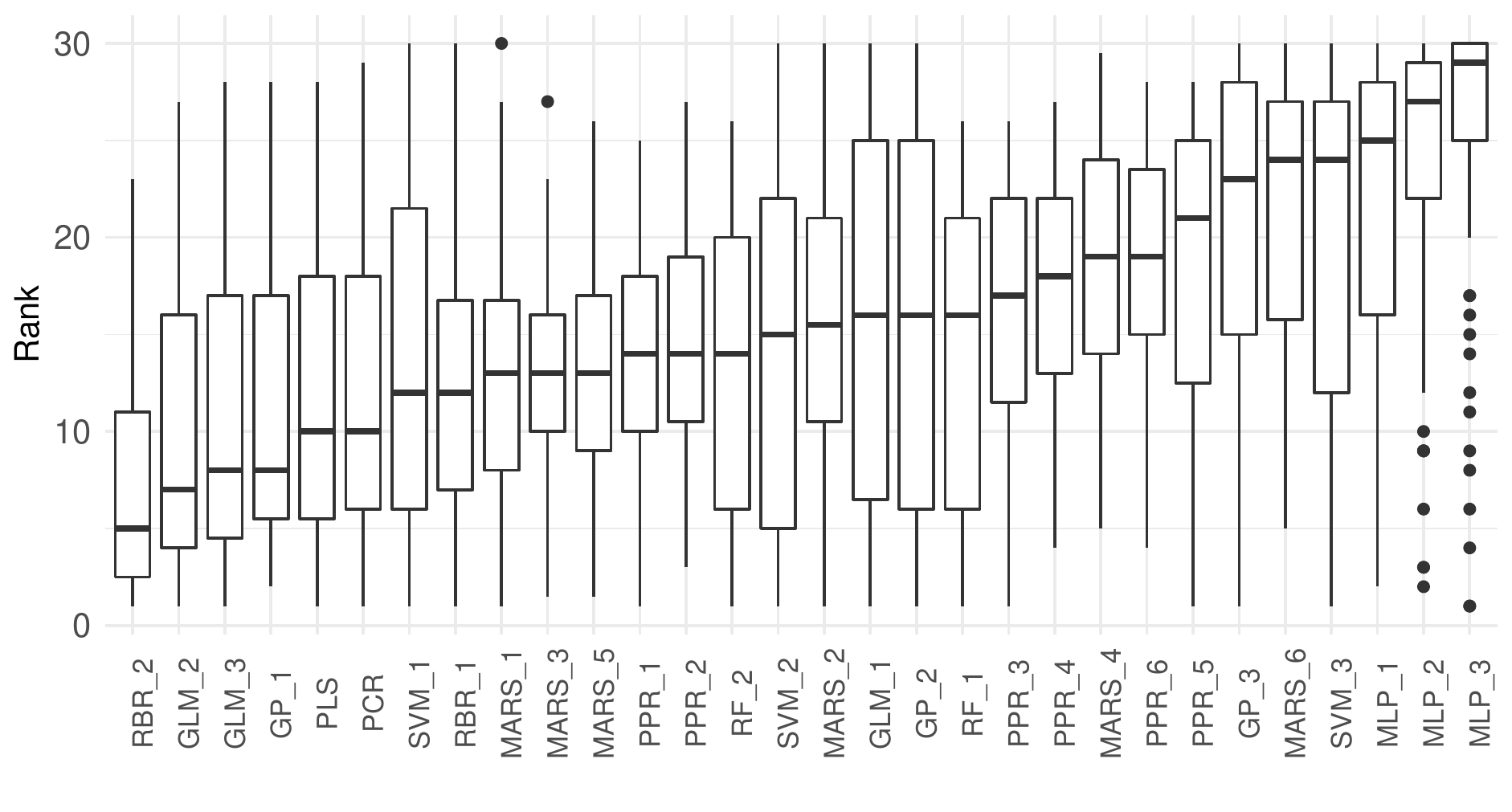}
\caption{Boxplots showing the distribution of the rank of each individual model in the ensemble across the 90 time series}
\label{fig:eda1}
\end{figure}

\begin{figure}[t]
\centering
\includegraphics[width=\textwidth]{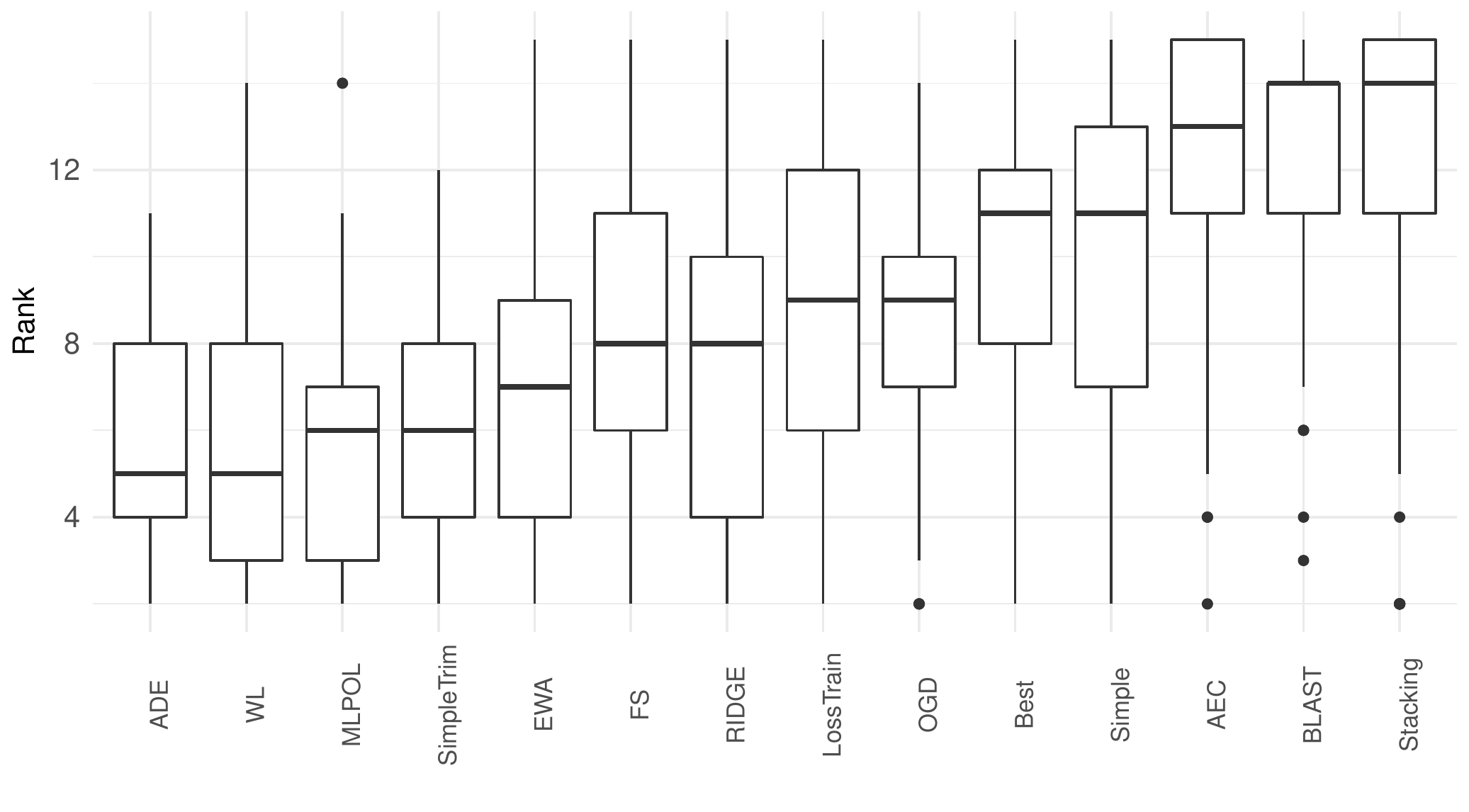}
\caption{Boxplots showing the distribution of the rank of each combination approach across the 90 time series}
\label{fig:eda2}
\end{figure}

In Figures \ref{fig:eda1} and \ref{fig:eda2}, we present an exploratory analysis of the results. The first figure show the distribution of the rank of each individual predictive model used in the ensembles across the 90 time series. The second figure presents a similar analysis for the forecast combination methods (which aggregate the individual models shown in Figure \ref{fig:eda1}).

The results of this analysis show that the \texttt{RBR\_2} model is the one with best median rank among the individual predictive models. Regarding the ensembles, \texttt{ADE} shows the best median rank. These results corroborate those from \cite{cerqueira2019arbitrage}. 
Figure \ref{fig:eda1} shows the practical implications of the No Free Lunch theorem. Essentially, no single approach is the best suited for all problems. This diversity in relative predictive performance is also visible in Figure \ref{fig:eda2}.


\subsection{Main Results}

In this section, we address each research question outlined in Section \ref{sec:rq} in turn.

\subsubsection{Q1 -- Students versus Teachers}

\begin{figure}[t]
\centering
\includegraphics[width=\textwidth]{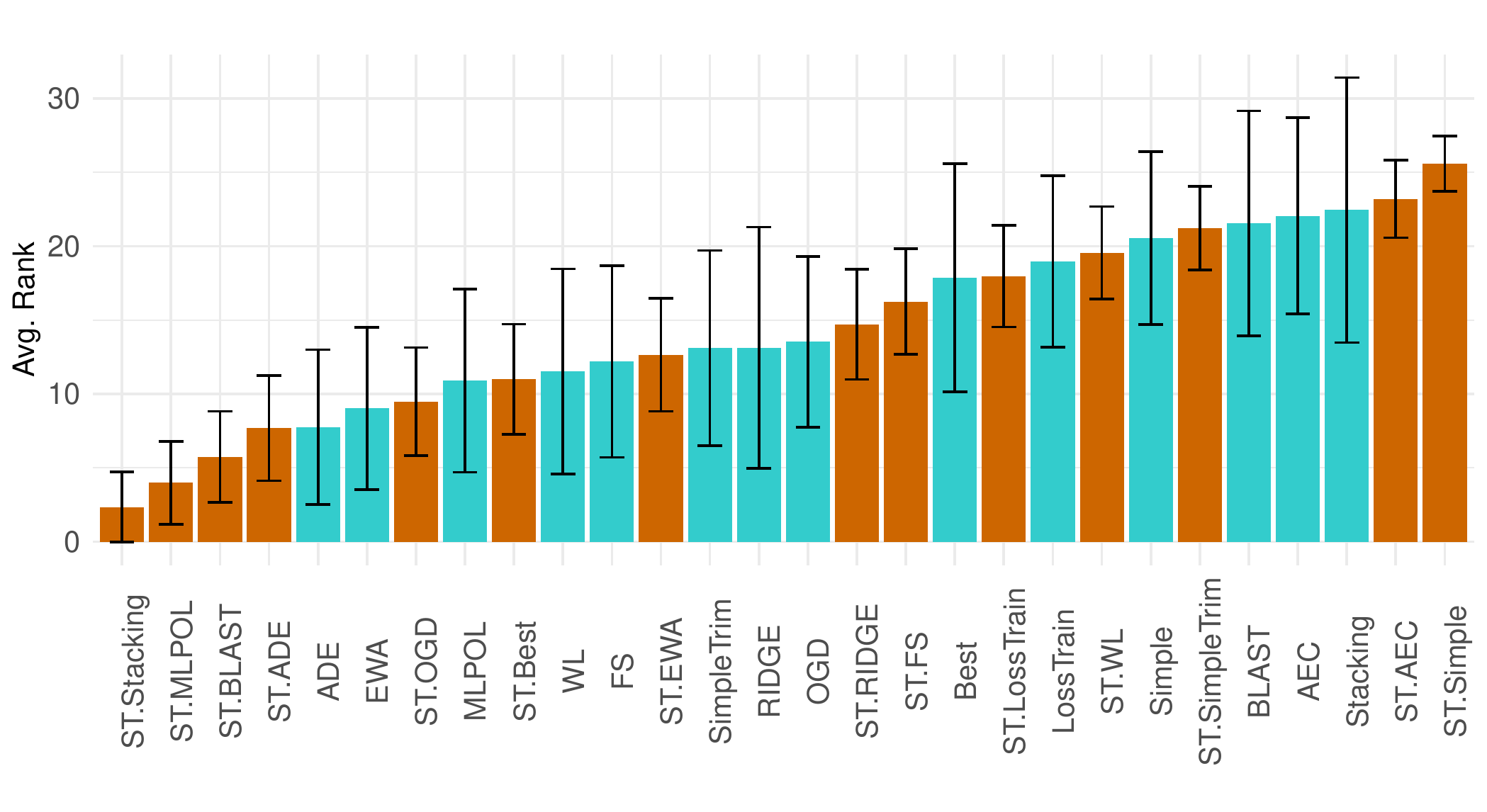}
\caption{Average rank, and respective standard deviation, of each method across the 90 time series. This analysis compares dynamic combination approaches with their respective variant applied using the ST training approach. The latter are denoted with a ``ST'' prefix.}
\label{fig:avgrank_all}
\end{figure}

\begin{figure}[t]
\centering
\includegraphics[width=\textwidth]{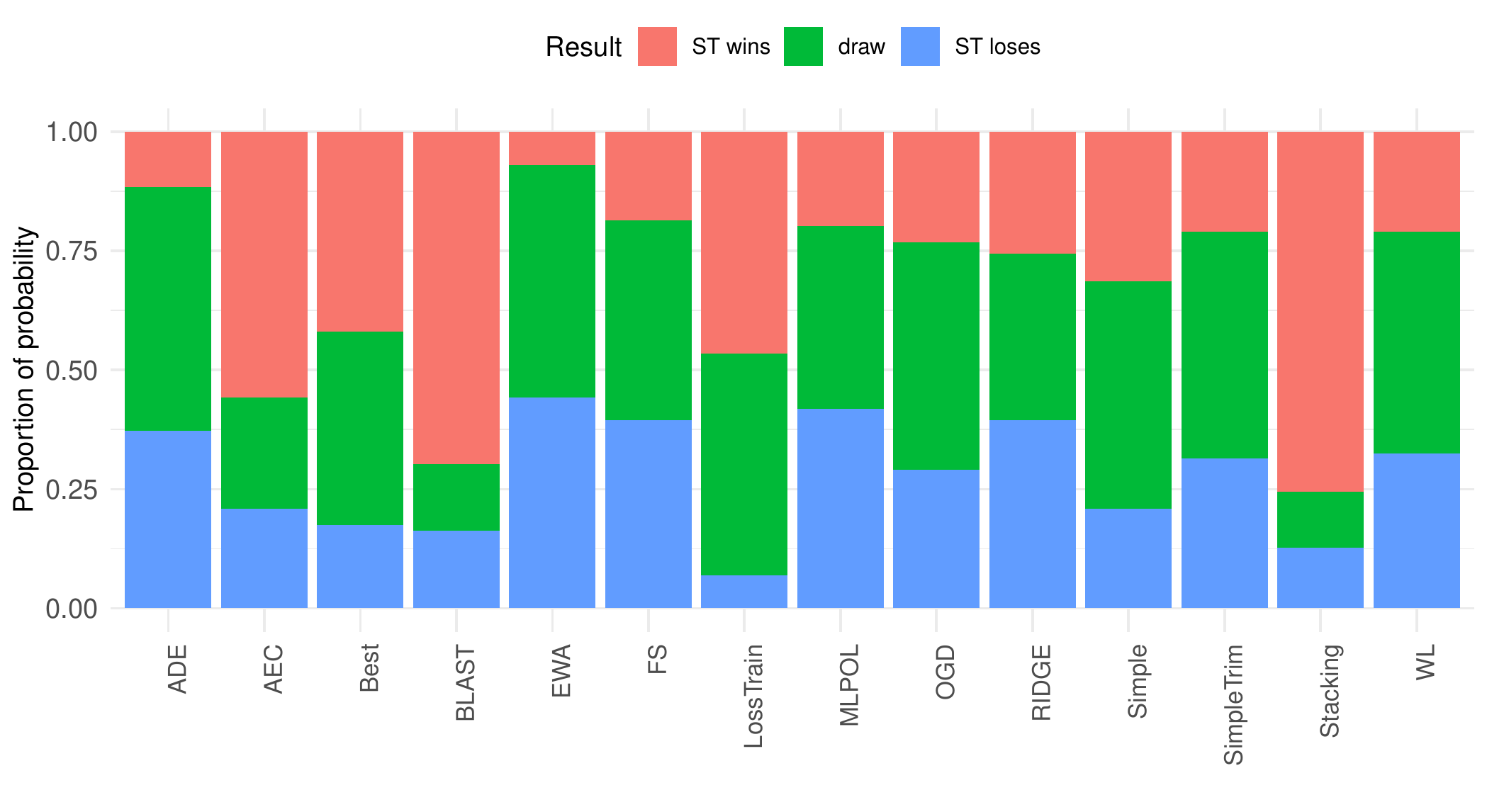}
\caption{Proportion of probability of the combination methods applied using ST winning/drawing/losing according to the Bayes sign test, for each ensemble method.}
\label{fig:bayes_comb}
\end{figure}

Figures \ref{fig:avgrank_all} and \ref{fig:bayes_comb} show the results that address the research question \textbf{Q1}. To recapitulate, \textbf{Q1} addresses the main question of this paper: How does the predictive performance of a compressed (student) model compare with ensemble methods (teachers)?

Figure \ref{fig:avgrank_all} shows the average rank, and respective standard deviation, of the ensemble methods and their compressed version. The latter is denoted with the prefix ``ST.'' and the respective bar is colored as orange.
In this analysis, we focus on a variant of the model tree as a student model (\texttt{RBR\_2}). The model tree, also known as M5, is a tree-based method proposed by \cite{quinlan1993combining}. This method extends a decision tree by including linear models in the decision nodes of the tree. This approach has been shown to provide a competitive predictive performance in forecasting problems by \cite{cerqueira2019arbitrage}. 
The results of this analysis are remarkable. 
The top four methods with best average rank (and six out of the top ten) are all student \texttt{RBR\_2} models, trained to emulate the behaviour of the respective combination approach. In particular, compressing the ensemble using \texttt{Stacking} as the teacher leads to the best average rank in this analysis. \texttt{ADE}, which is fifth overall, presents the best score among the dynamic ensemble approaches. Finally, it is also noticeable that the student models present a smaller standard deviation of their average rank score.

Some of the teachers of the methods showing the best score in average rank (\texttt{Stacking}, \texttt{BLAST}) do not show a good relative performance when compared to other dynamic ensemble methods (e.g. \texttt{ADE} or \texttt{EWA}). This suggests that better combination functions do not necessarily lead to better teachers.

We apply a Bayesian analysis to assess the statistical significance of the results. We employed the Bayes sign test to compare pairs of methods across multiple problems. We define the \textit{region of practical equivalence} \citep{benavoli2017time} (ROPE) to be the interval [-1\%, 1\%]. Essentially, this means that two methods show indistinguishable performance if the difference in percentage difference in predictive performance between them falls within this interval. For a thorough read on Bayesian analysis for comparing predictive models we refer to the work by \cite{benavoli2017time}.

Figure \ref{fig:bayes_comb} complements the previous figure by breaking the analysis per ensemble method with the described Bayes Sign test. This figure illustrates the proportion of probability that the methods applied using the \texttt{ST} training approach win, draw (result within the ROPE), or lose, for each respective combination. 
Taking \texttt{Simple} as an example, the probability that, when using model compression, the student model wins significantly is about 30\%. The probability of winning significantly is about 20\%. The probability of a draw is the remaining 50\%. 
Overall, the results are quite even. Notwithstanding, it is clear that some ensemble methods \textit{react better} (e.g. \texttt{Stacking}, \texttt{BLAST}) to the application of model compression than others (e.g. \texttt{EWA}).

\begin{figure}[t]
\centering
\includegraphics[width=\textwidth]{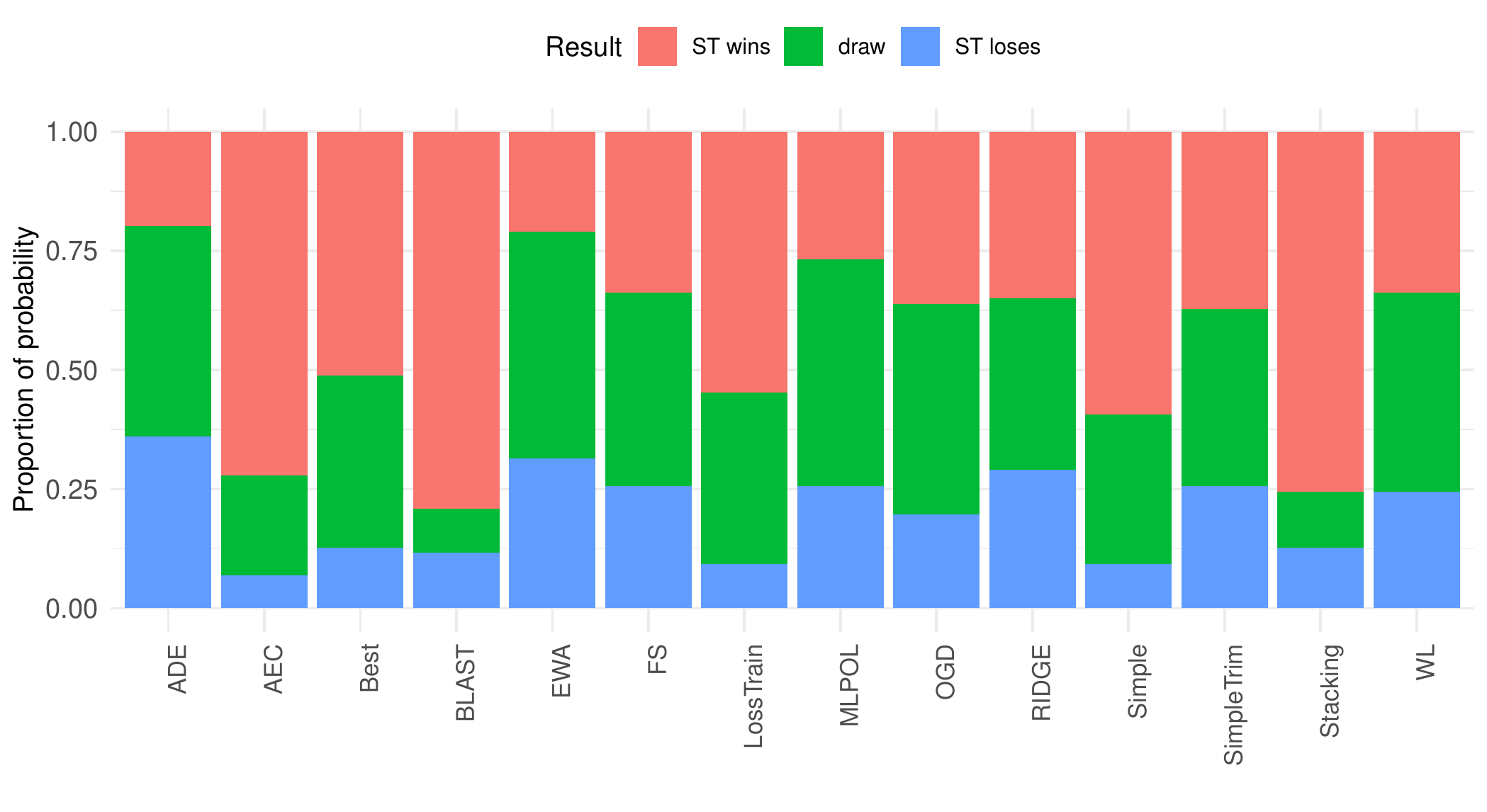}
\caption{Proportion of probability of the \texttt{ST.Stacking} student model winning/drawing/losing against each ensemble method according to the Bayes sign test.}
\label{fig:bayes_comb2}
\end{figure}

Finally, Figure \ref{fig:bayes_comb2} presents a similar analysis to Figure \ref{fig:bayes_comb}. We compare each teacher against the student \texttt{ST.Stacking}, which showed the best average rank in Figure \ref{fig:avgrank_all}. The results are comparable. However, when comparing \texttt{ST.Stacking} with the best performing ensemble (e.g. \texttt{ADE}, \texttt{EWA}), the probability of the latter winning significantly is larger than the opposite. These results may seem paradoxical. \cite{benavoli2016should} points out that this issue arises because, when analysing average ranks, the comparison between two models depends on the performance on the remaining pool of models. Therefore, the Bayes sign test provides a more reliable comparison between pair of models.

\subsubsection{Q2 \& Q3 -- Individual Models With and Without Compression}

Research questions \textbf{Q2} and \textbf{Q3} address the comparison of individual models training with the original target variable and individual models training by model compression. In this analysis, we focus on the \texttt{Stacking} aggregation method as a teacher, in which predictive models are combined according to the arithmetic mean.

\begin{figure}[t]
\centering
\includegraphics[width=\textwidth]{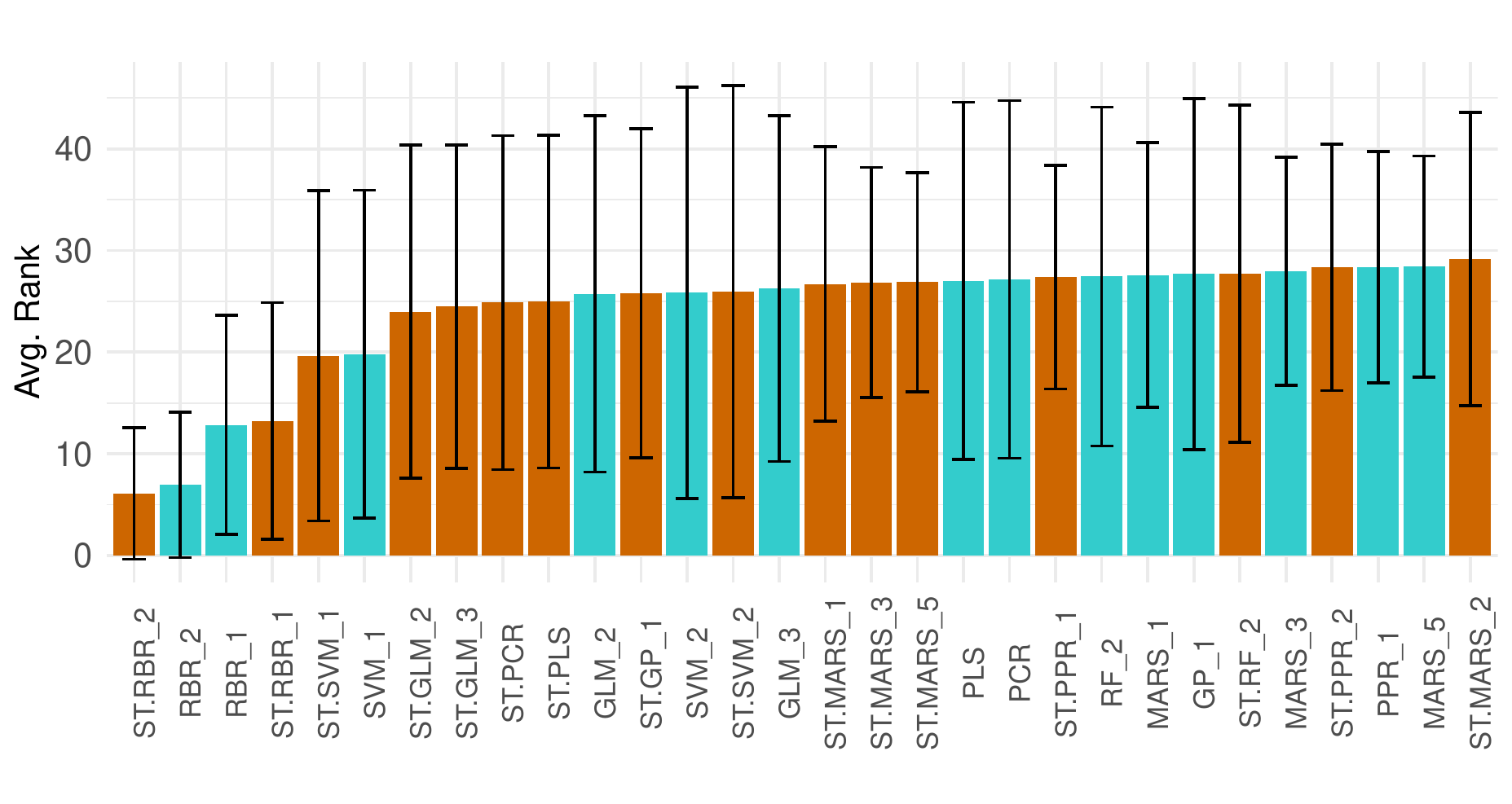}
\caption{Average rank, and respective standard deviation, of each individual method trained using the original data and trained by compressing the \texttt{Stacking} ensemble method. The latter are denoted using the ``ST'' prefix.}
\label{fig:avgrank_indcompressed}
\end{figure}

We start this analysis by showing the average rank of each method in Figure \ref{fig:avgrank_indcompressed}. The study includes the 30 individual models trained using the original data plus their respective variant trained according to the ST approach. Similarly to Figure \ref{fig:avgrank_all}, methods trained using the ST approach are denoted with a ``ST.'' prefix and colored as orange (as opposed to light-blue). The labels of the figure only show the ID of each method (according to Table \ref{tab:expertsspecs}). Finally, in the interest of conciseness, we show only the 30 methods with best average rank. 

The results show that the methods with better average rank are mostly student models. The rule-based regression methods presents the best predictive performance (both versions of it), which lead to its choice in the analysis presented in the previous section.

\begin{figure}[t]
\centering
\includegraphics[width=\textwidth]{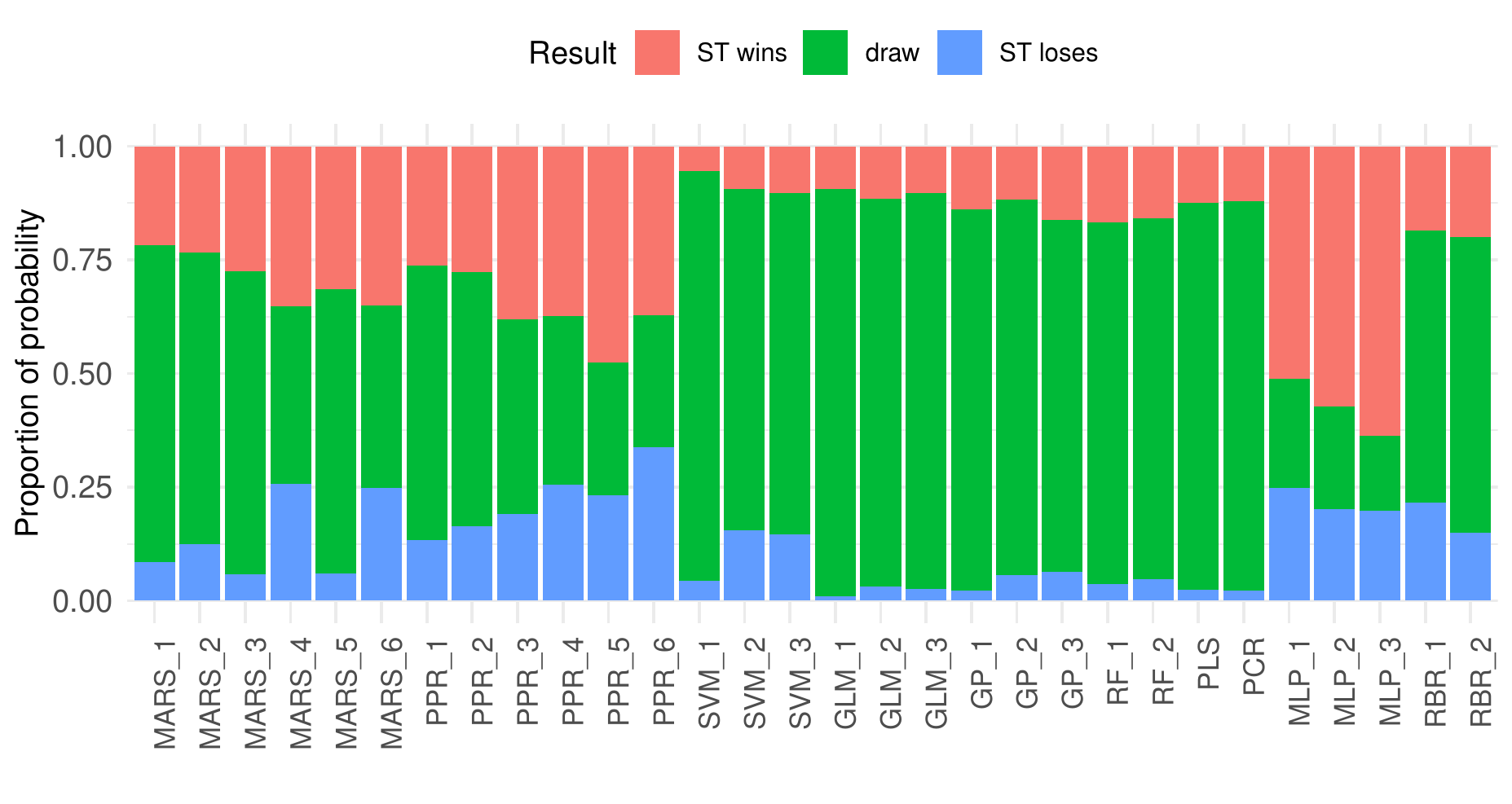}
\caption{Proportion of probability of model compression winning/drawing/losing according to the Bayes sign test, for each individual method.}
\label{fig:bayessign_ind}
\end{figure}

Similarly to before, we complement the average rank analysis with a Bayesian study. Figure \ref{fig:bayessign_ind} shows the results of this experiment, which follow an identical rationale as Figure \ref{fig:bayes_comb}. For the majority of methods, the probability of the model trained according to the ST approach winning significantly is considerably larger than the opposite.


In summary, the results suggest that training a given learning algorithm with the predictions of the ensemble method \texttt{Stacking} leads to, on average, a better predictive performance when compared to the same learning algorithm trained with the original target variable. Following the results of the previous section, these results were expected. The models under comparison in the section are single methods, while in the previous section we tested the predictive performance of ensemble approaches.


\subsubsection{Q4 -- Students versus Traditional Forecasting Methods}

\textbf{Q4} is related to the comparison of the proposed methods with traditional forecasting models, such as \texttt{ARIMA}, \texttt{ETS}, and \texttt{TBATS} \cite{hyndman2018forecasting}. Recently, Makridakis \cite{makridakis2018statistical} presented evidence that simple statistical methods, including a simple random walk, systematically outperform machine learning methods for time series forecasting. \cite{cerqueira2019machine} suggested that these results are only valid due to the small training sample size used by the authors. Notwithstanding, in the interest of completeness, we test these methods in this work.

\begin{figure}[t]
\centering
\includegraphics[width=\textwidth]{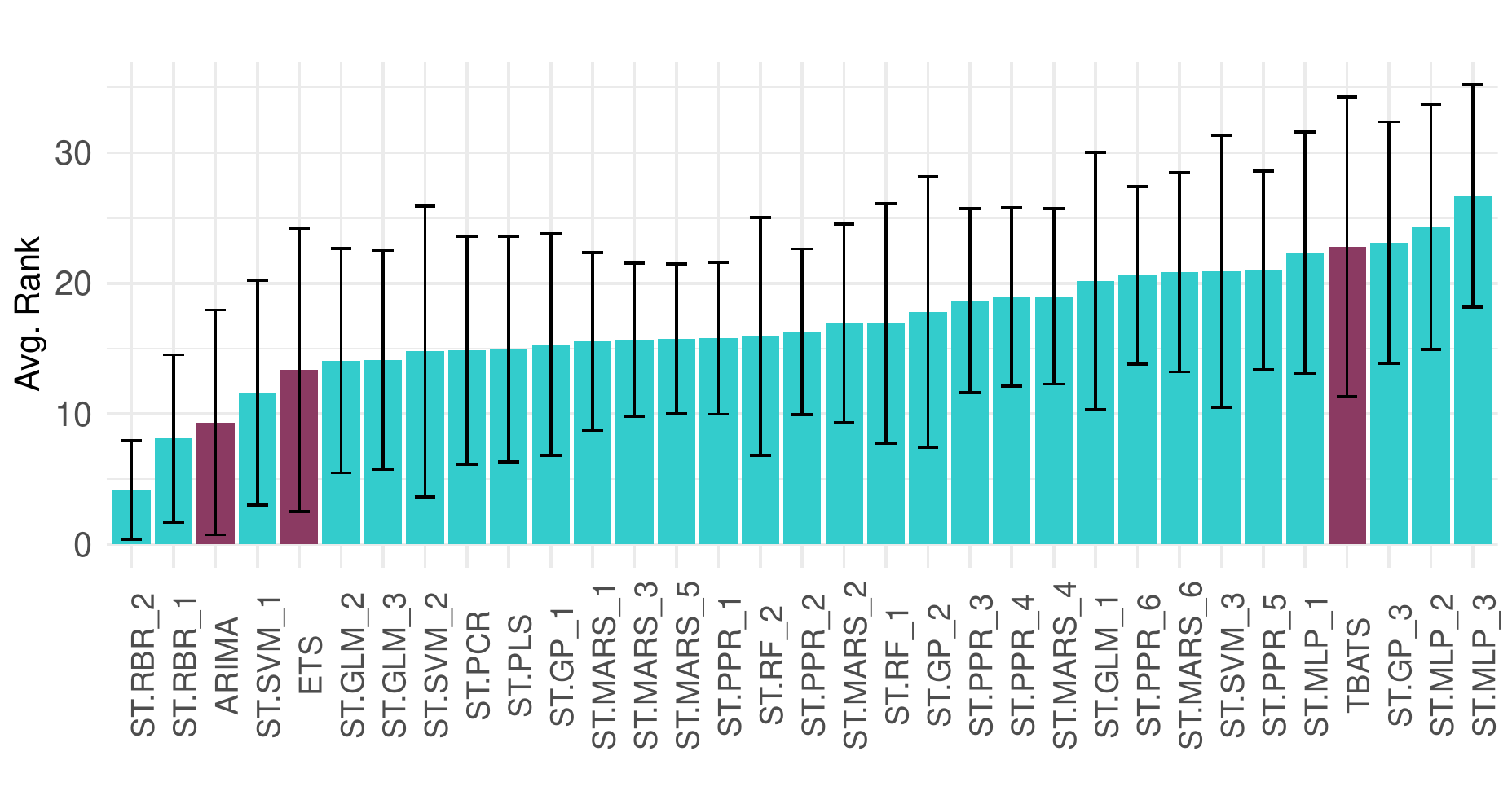}
\caption{Average rank, and respective standard deviation, of each student method (using \texttt{Simple} as teacher) and three automatic traditional forecasting approaches: \texttt{ARIMA}, \texttt{ETS}, and \texttt{TBATS}.}
\label{fig:avgrank_trad}
\end{figure}

\begin{figure}[t]
\centering
\includegraphics[width=\textwidth]{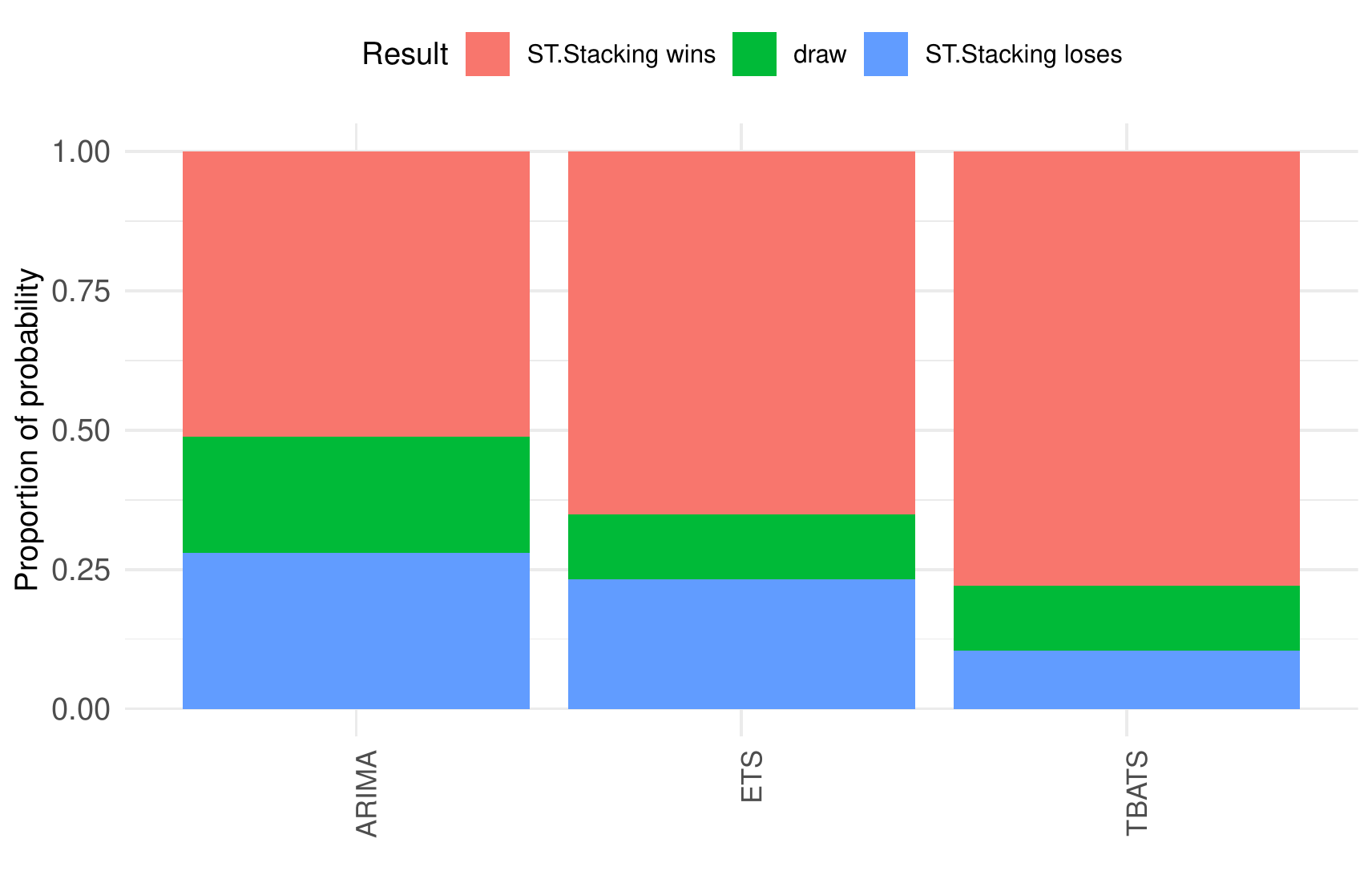}
\caption{Proportion of probability of \texttt{ST.Stacking} winning/drawing/losing according to the Bayes sign test against three traditional forecasting approaches: \texttt{ARIMA}, \texttt{ETS}, and \texttt{TBATS}.}
\label{fig:bayes_trad}
\end{figure}

The results are shown in Figures \ref{fig:avgrank_trad} and \ref{fig:bayes_trad}. The first illustrates the average rank, and respective standard deviation of each student individual model with \texttt{Stacking} as teacher and three automatic traditional forecasting approaches: \texttt{ARIMA}, \texttt{ETS}, and \texttt{TBATS}. The latter are color-coded as pink, while the others are colored as orange as before. \texttt{ARIMA} shows the best average rank among the traditional methods, and ranks third overall. These results suggest that it may be beneficial to include these methods in the pool of models composing the ensemble. To assess the significance of the results we carried a Bayesian analysis identical to the previous ones. We compared \texttt{ST.Stacking} (trained using a \texttt{RBR\_2} method) with each automatic traditional forecasting method. The results show that \texttt{ST.Stacking} has a larger probability of winning significantly against each approach.

\subsubsection{Q5 -- Teaching Data}

We now address \textbf{Q5}, which is related to which data is more appropriate to use for training the student models. \cite{bucilua2006model} presented a method for generating synthetic instances to this effect in classification problems. However, \cite{hinton2015distilling} suggests that we can use the predictions of the teachers in the training data. This may sound counter-intuitive because  predicting in the same data used to train the models is accepted to lead to biased results. Motivated by this belief, \cite{cerqueira2019arbitrage} used a blocked prequential approach to retrieve unbiased predictions from the training set to train a meta-learning layer of models.

\begin{figure}[t]
\centering
\includegraphics[width=\textwidth]{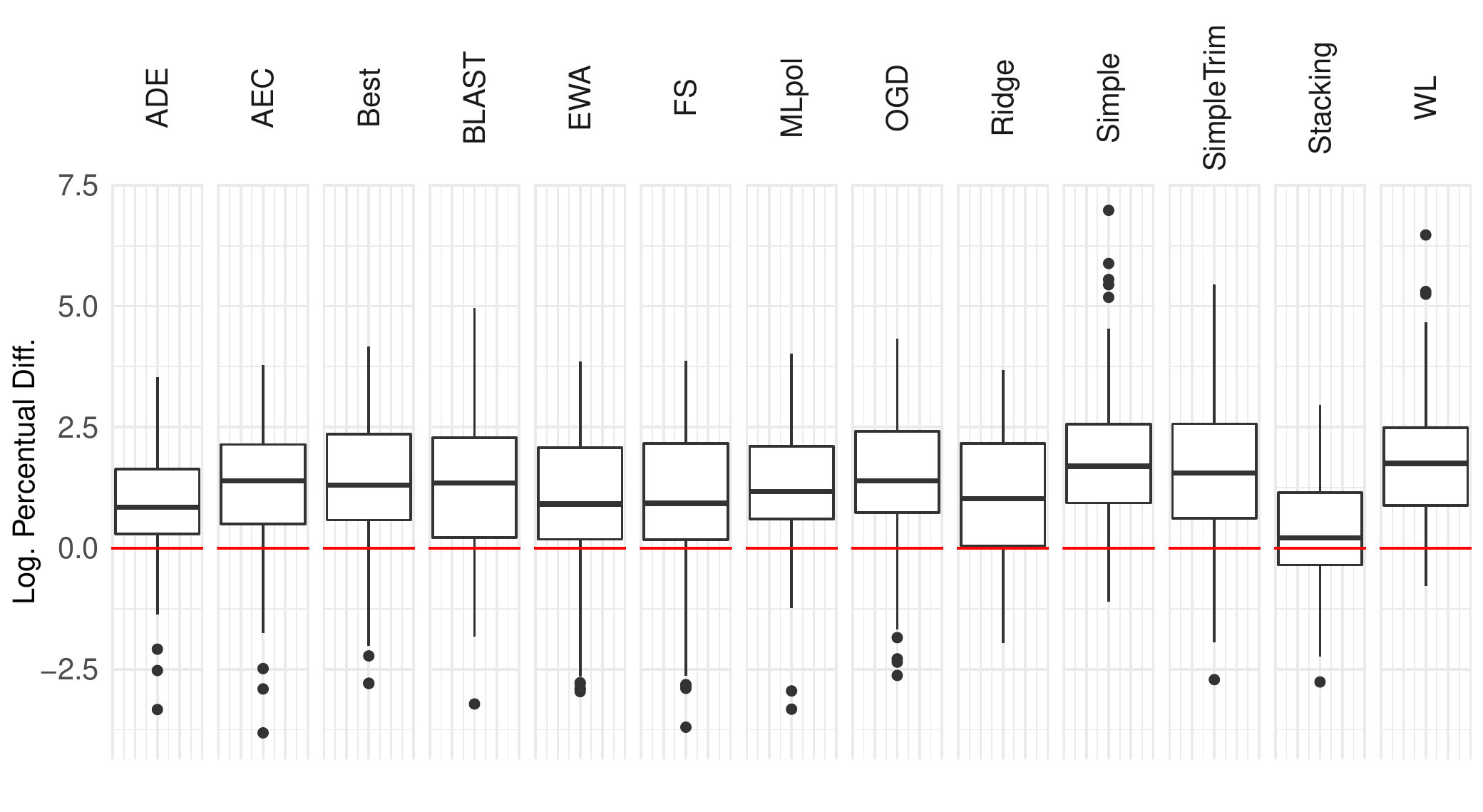}
\caption{Distribution of percentual difference (log scaled) in MASE between (i) the \texttt{RBR\_2} student model trained using the predictions of the teacher in the training data; and (ii) the same method trained using out-of-bag predictions of the teacher, retrieved using a blocked prequential approach. This analysis is carried for each combination approach. Positive values denote better performance for the first approach (i).}
\label{fig:pd_trvld}
\end{figure}

In this section, we test the following two alternatives: 
\begin{description}
    \item[a)] use the predictions of the teachers in the data used to train them to train the student models;
    \item[b)] use a blocked prequential approach to retrieve unbiased predictions from the training set to this effect. 
\end{description}

\noindent Figure \ref{fig:pd_trvld} shows the results of this analysis. 
It shows the distribution of the percentage difference in MASE (log scaled) between the \texttt{RBR\_2} student model trained using (a) and the student models trained according to (b), for each teacher. Positive values denote better performance by the first alternative (a). The results clearly show that alternative (a) leads to a better predictive performance for all teacher models.

\subsubsection{Q6 -- Computational Costs}

Finally, we analyse the relative difference in computational costs between students and teachers (\textbf{Q6}). We present the results in relative terms since these results may vary due to different machine specifications. We focus on the \texttt{RBR\_2} student model, which presented the best relative predictive performance. In the interest of conciseness, we also focus on the \texttt{Simple} method as the teacher. We split the comparison (between \texttt{ST.Simple} and \texttt{Simple}) in two dimensions: the time it takes to predict the test instances, and the space required to store the predictive model (or set of predictive models for \texttt{Simple}) in memory. On top of this, we also study the space required to store \texttt{ST.Simple} (a \texttt{RBR\_2} model).

\begin{figure}[t]
\centering
\includegraphics[width=\textwidth]{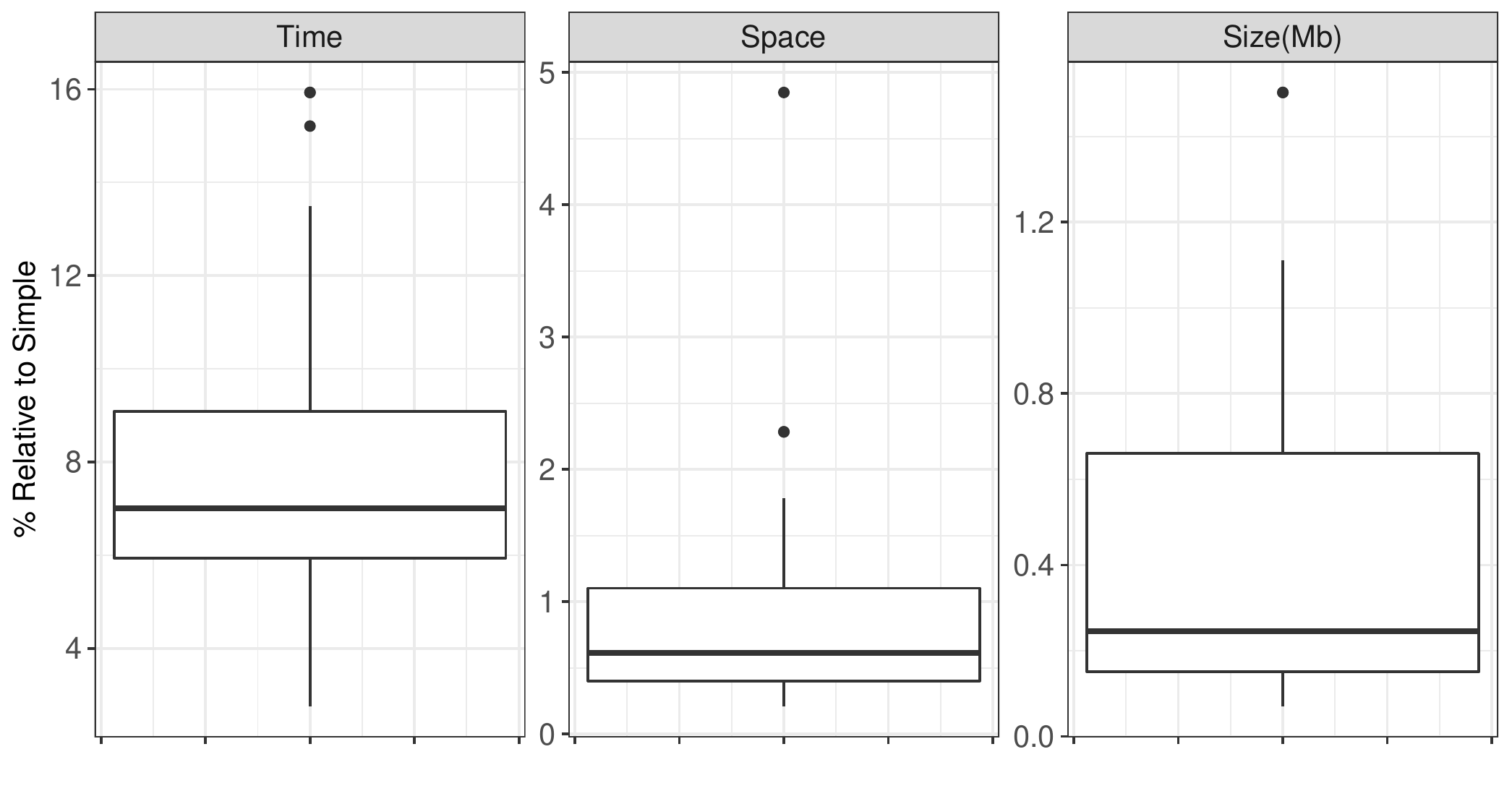}
\caption{Relative computational costs between \texttt{Simple} and \texttt{ST.Simple} (an \texttt{RBR\_2} model): \textit{Time} necessary to predict all test instances; \textit{Space} required for storing the predictive model; and \textit{Size} of the student model, in megabytes.}
\label{fig:cc}
\end{figure}

The results are shown in Figure \ref{fig:cc}. The figure illustrates the distribution of the computational costs of \texttt{ST.Simple} relative to \texttt{Simple}, across the 90 time series problems. On average, \texttt{ST.Simple} takes less than 8\% of the time \texttt{Simple} takes to predict the test observations. Moreover, it requires less than 1\% of the space of the ensemble (on average). The median space required to store the \texttt{RBR\_2} model representing \texttt{ST.Simple} is less than 0.4 megabytes.

The \texttt{Simple} aggregation approach is the most basic one, combining the available predictions with a simple arithmetic mean. In effect, the results of this analysis would magnify in favor of student models for different forecast combination approaches. This is due to the extra hurdle of having to frequently compute and store the weights of each individual model in the ensemble.

\section{Discussion}\label{sec:discussion}

\subsection{Main Results}

In the previous section we showed two important results with practical implications for the forecasting community. 
First and foremost, it is possible to compress a dynamic forecasting ensemble into a compact individual model and retain a competitive predictive performance. Students show a better average rank relative to teacher, but the pairwise significance analyses showed comparable results. However, these are accompanied with a drastic reduction in computational costs in favor of student models.
Second, a given individual method performs significantly better when trained according to the ST approach, as opposed to being trained using the original target variable.
Similar results to these have been reported in different predictive tasks, and different individual models and combination approaches \citep{bucilua2006model,hinton2015distilling}.

The goal of a predictive model is to generalize well to unseen observations. \cite{hinton2015distilling} suggests that model compression encourages this by having a learning algorithm mimicking the function of an ensemble, which typically generalizes well, as opposed to having the algorithm optimizing performance on training data.

\subsection{Future Work}

We showed the benefits of model compression in time series forecasting problems using an extensive experimental design. We believe these results represent a significant contribution to the field. Particularly embedded systems, where predictive models in small edge devices must be as compact as possible. As future work, it would be interesting to analyse the impact of the proposed methodology in a particular case study within this topic, for example in terms of energy efficiency.

Time-dependent data is prone to concept drift. Often, the non-stationarities present in time-series are re-occurring due to, for example, seasonality. However, sometimes the underlying concept drift to another one unknown hitherto. In this context, an interesting research line would be to analyse to application of model compression in such scenarios. That is, study how model compression approaches behave under concept drift. A standard approach would be to monitor the error of the deployed model using state of the art concept drift detection and adaptation approaches \citep{gama2014survey}. 

\section{Conclusions}

Model compression \citep{bucilua2006model,hinton2015distilling} has enabled the transfer of the capacity of large ensembles into compact predictive models. In this work, we show the applicability of this type of approaches in time series forecasting problems. 
We show that forecast combination approaches can be compressed into an individual predictive model. This individual model retains a competitive predictive performance with the ensemble while being faster and more compact. The results shown in this paper can have an important impact in two different research topics involving forecasting: resource-aware environments, such as embedded systems related to the \textit{internet-of-things}, and sensitive domains of application which require accurate yet transparent models.

\section*{Acknowledgments}
The authors would like to acknowledge the valuable input of anonymous reviewers. 

\bibliographystyle{model5-names}
\biboptions{authoryear}

\end{document}